\documentclass[10pt,twocolumn,letterpaper]{article}

\usepackage[accsupp]{axessibility}  
\usepackage{iccv}
\usepackage{times}
\usepackage{epsfig}
\usepackage{graphicx}
\usepackage{amsmath}
\usepackage{amssymb}

\usepackage[font=small,labelfont=bf,tableposition=top]{caption}
\usepackage{blindtext}
\usepackage{capt-of,etoolbox}

\usepackage{multicol}
\usepackage{multirow} 
\usepackage{makecell} 
\usepackage{setspace}
\usepackage{mathrsfs}  
\usepackage[ruled,vlined]{algorithm2e}
\usepackage{subcaption}
\usepackage{footmisc} 
\usepackage{flafter}
\usepackage{afterpage}
\usepackage{array}
\newcolumntype{L}[1]{>{\raggedright\let\newline\\\arraybackslash\hspace{0pt}}m{#1}}
\newcolumntype{C}[1]{>{\centering\let\newline\\\arraybackslash\hspace{0pt}}m{#1}}
\newcolumntype{R}[1]{>{\raggedleft\let\newline\\\arraybackslash\hspace{0pt}}m{#1}}

\usepackage[pagebackref=true,breaklinks=true,letterpaper=true,colorlinks,bookmarks=false]{hyperref}

\iccvfinalcopy 

\def\iccvPaperID{12492} 

\ificcvfinal\fi

\newcommand{\insertfig}{{\includegraphics[width=\linewidth]{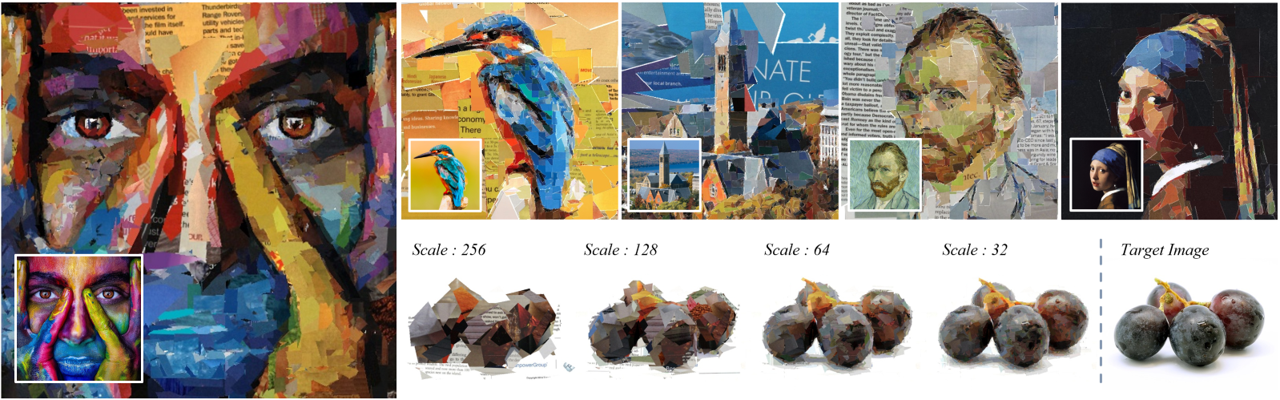}}
\begin{flushleft}\refstepcounter{figure}\textbf{Figure~\thefigure}: The results of neural collage transfer. Given an image and materials, each output collage was generated based on \\ the proposed complexity-aware multi-scale collage method. The sequence of grapes shows our collage generation process.\end{flushleft}}

\makeatletter
\apptocmd{\@maketitle}{\centering\insertfig}{}{}
\makeatother

\begin{document}

\title{Neural Collage Transfer: Artistic Reconstruction via Material Manipulation}

\author{Ganghun Lee, Minji Kim, Yunsu Lee, Minsu Lee\thanks{Corresponding authors}, Byoung-Tak Zhang\footnotemark[1]\\
Seoul National University\\
{\{\tt\small khlee, mjkim, yslee, mslee, btzhang\}@bi.snu.ac.kr}
}

\maketitle

\ificcvfinal\thispagestyle{empty}\fi

\begin{abstract}
Collage is a creative art form that uses diverse material scraps as a base unit to compose a single image.
Although pixel-wise generation techniques can reproduce a target image in collage style, it is not a suitable method due to the solid stroke-by-stroke nature of the collage form.
While some previous works for stroke-based rendering produced decent sketches and paintings, collages have received much less attention in research despite their popularity as a style.
In this paper, we propose a method for learning to make collages via reinforcement learning without the need for demonstrations or collage artwork data.
We design the collage Markov Decision Process (MDP), which allows the agent to handle various materials and propose a model-based soft actor-critic to mitigate the agent's training burden derived from the sophisticated dynamics of collage.
Moreover, we devise additional techniques such as active material selection and complexity-based multi-scale collage to handle target images at any size and enhance the results' aesthetics by placing relatively more scraps in areas of high complexity.
Experimental results show that the trained agent appropriately selected and pasted materials to regenerate the target image into a collage and obtained a higher evaluation score on content and style than pixel-wise generation methods. Code is available at \url{https://github.com/northadventure/CollageRL}.
\vspace{-2mm}  
\end{abstract}

\section{Introduction}
Collage, derived from the French word \textit{coller} meaning \textit{to glue}, is a fundamental art form where disparate scraps of images are assembled and arranged geometrically to create a complete scene.
The intriguing and alluring impression of collage was influenced by the art movement of \textit{Cubism}, particularly the work of \textit{Pablo Picasso}, and has now become a prevalent art style.
However, creating high-quality collage artworks in the style of artists such as \textit{Derek Gores} requires professional-level skills, much like other art forms such as painting.
Collage may seem like a simple variation of painting, but it is much more challenging when attempted by artificial agent for several reasons.
Unlike predefined brush strokes in paintings, the strokes of a collage depend on selecting materials from a diverse range of candidates, which involves uncertainty.
After the material selection, the complicated manipulation process of cutting and pasting follows.
Moreover, detailed representations are strictly limited when the content of material pieces is unmodifiable.
Even for a human, a large amount of experience is required to handle these difficulties.

Fortunately, recent advances in AI-based art creation have made art more accessible to ordinary people. 
Pixel-based generation approaches evolved from Deep Convolutional Neural Networks (DCNNs) \cite{vgg, resnet}, Variational Auto-Encoders (VAEs) \cite{vae}, and Generative Adversarial Networks (GANs) \cite{gan} have fostered the creation of novel artworks \cite{deepdream, aican} and transferred images into various styles \cite{gatys, nstnet, pix2pix, cyclegan, drbgan, adain, adaattn, clipstyler} under Neural Style Transfer (NST) \cite{nst}.
However, pixel-based approaches directly discover the final look in the pixel space; the result can be seen as unnatural and cannot obtain a natural creation sequence.

Since most art formulations enjoy stroke-by-stroke, a more natural and aesthetic collage can be obtained if a machine learns to create collages in a stroke-by-stroke fashion.
There have been some previous learning approaches in stroke-based rendering (SBR) \cite{sbr} to imitate stroke-by-stroke image generation, such as supervised learning \cite{sketchrnn, painttransformer, sketchbert, doodleformer}, reinforcement learning (RL) \cite{spiral, learning2paint}, and optimization \cite{stylizedneuralpainting, parameterizedbrushnst, clip-clop}.
Although they produced attractive sketches and paintings, they are partly incompatible with the concept of collages.
For supervised learning, public collage process data is hardly available, and generating sufficient data is almost infeasible.
Existing RL and optimization methods assuming the fixed material structure for every stroke do not align well with collages incorporating dynamic materials.
Overall, collage has received comparatively less attention in the artistic creation domain.

In this paper, we introduce a novel RL-based method for learning how to generate a collage artwork using a given target image and materials.
RL offers experience-based learning to handle various possible situations in the collage procedure without data cost, leading to the development of distinct creation styles.
To effectively train the agent, we build a novel environment, the collage Markov Decision Process (MDP), which allows the agent to explore a wide range of material spaces to learn appropriate material selection and manipulation skills.
We propose the model-based Soft Actor-Critic (MB-SAC) to mitigate the burden of predicting complicated dynamics in collage.
At each training step, the agent observes the canvas, the given material, and the target image to decide on cutout shapes for the material and the pasting locations on the canvas.
The reward increases as the canvas becomes more similar to the target image, encouraging the agent to learn more elaborate behaviors to reproduce the original content in a collage form.
Moreover, we extend the trained collage agent to make a multi-scale collage for any size of the target images, considering the image complexity for more aesthetic appeal.

Our extensive experimental results show that our proposed method enables an agent to learn the nature of collage autonomously without expensive collage datasets or collage demonstrations.
Moreover, the proposed method can handle target images of any size, and the generated collage produces sophisticated and aesthetic output.
This is because the agent can select appropriate scraps from materials and place more scraps in areas of high complexity.
Furthermore, through quantitative analysis, we have shown that our approach outperforms other competitive methods in terms of image similarity from the target image and semantic consistency using CLIP \cite{CLIP}-based measures and user study.
These results demonstrate that our method successfully transfers the target image to collage style while preserving its content.


\begin{figure*}[t]
\begin{center}
\includegraphics[width=1\linewidth]{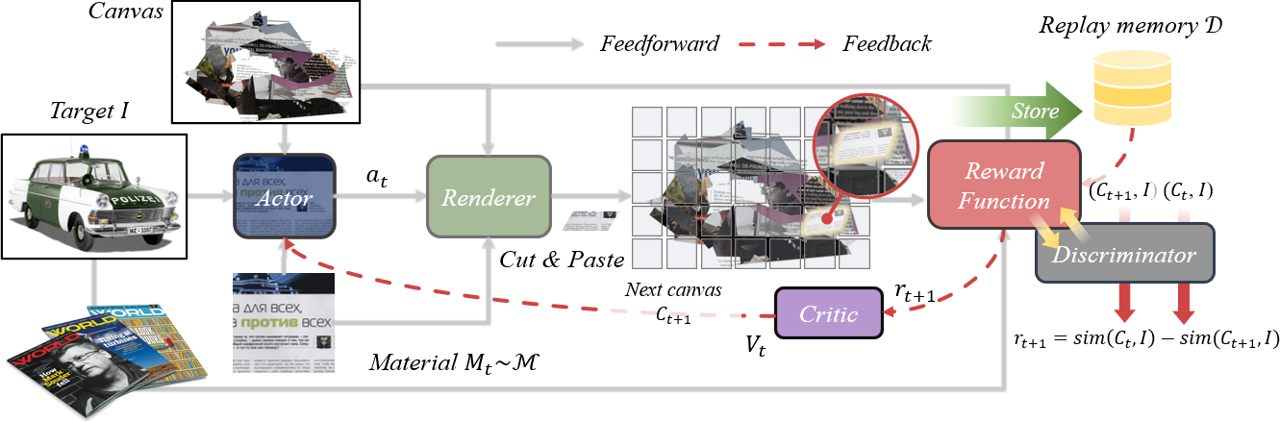}
\end{center}
\caption{Overview of collage MDP and its associated training process. The agent first observes an image triplet consisting of a canvas, a target image, and a material image and then makes a scrap of the material that will make the canvas fit the target image as closely as possible. Once the scrap has been pasted onto the canvas, the transition process is saved in replay memory for training later. During the update stage, the discriminator in the reward function is updated using samples from the replay memory. The actor and critic are then updated using the rewards obtained from these samples.}
\label{fig:overview}
\end{figure*}

\section{Related Work}

\subsection{Neural Style Transfer}
For the past couple of years, Neural Style Transfer (NST) \cite{nst} has been the standout tool in artistic style transfer.
NST aims to convert a target image to the given target style preserving the content of the target image.
General approaches of NST rely on pixel-wise gradient descent which is directly applied onto the converted image \cite{gatys}, or on using models trained to approximate the target style distribution \cite{nstnet, adain, pix2pix, cyclegan, drbgan, p2gan}.
Some recent works support style transfer from text prompts \cite{clipstyler, stablediffusion}.
Although advanced NST models cover a wide range of styles, their utility for collage style is limited since the original purpose of NST may not match with collage.
The purpose of their pixel-wise style extraction is to look for patterns commonly seen for each style type.
However, as collages are composed of separate images, common patterns are rarely detected except for the edges between image pieces, thus not aligning with the purpose of NST.
This unique feature of collage demands another approach for collage style transfer.

\subsection{Stroke-based Rendering}
Stroke-based Rendering (SBR) is a non-pixel-based automated approach for generating non-photorealistic imagery by arranging discrete elements such as paint strokes or stipples \cite{sbr}.
Deep learning-based SBR was first proposed in Sketch-RNN \cite{sketchrnn} where human sketch demonstration is provided to train the Recurrent Neural Networks (RNNs).
However, supervised methods suffer from severe data collection and refinement costs.
Instead, methods using RL \cite{spiral, learning2paint} train the painting agent without supervision.
The gradient-based procedure optimization methods \cite{parameterizedbrushnst, stylizedneuralpainting} leverages the design of a fully-differentiable painting.
Studies using transformers \cite{transformer, doodleformer, sketchbert, painttransformer} or robotics applications \cite{scratch2sketch, sbrforrobotart} were also conducted.

Unfortunately, existing SBR-based research has mainly focused on the environments in which stroke structure is pre-modeled (\eg, sketches and paintings). 
These differ from collage environments where the available strokes (materials) are only known during the process.
In this paper, we extend \cite{learning2paint} to the collage domain, proposing a novel training environment and an alternative training algorithm.

\subsection{Collage Generation}
Several studies have explored collage generation, although not specifically collage transfer. Picture collage \cite{picturecollage2006, picturecollage2009, picturecollage2017} aims to assemble a set of given pictures into a complete image, while CLIP-CLOP \cite{clip-clop} generates collage artworks from text prompts using predefined strokes with modifiable properties. In contrast, our method uses non-predefined materials without unrealistic transformations for image transfer. Additionally, while their goal is text-based generation, our approach focuses on reproducing target images as natural collages. To the best of our knowledge, no research has been conducted on achieving this specific goal.


\section{Proposed Methods}
\subsection{Collage MDP}
\textbf{Preliminary.}
The problem of RL is formulated as a Markov Decision Process (MDP) consisting of a tuple $\langle \mathcal{S}, \mathcal{A}, \mathcal{P}, \mathcal{R}, \gamma \rangle$, where $\mathcal{S}$ is state space, $\mathcal{A}$ is action space, $\mathcal{P}:\mathcal{S}\times\mathcal{A}\rightarrow\mathcal{S}$ is the transition function, $\mathcal{R}:\mathcal{S}\times\mathcal{A}\rightarrow\mathbb{R}$ is the reward function, and $\gamma\in\left[0,1\right]$ is the discount factor.
The MDP has a state $s_t\in \mathcal{S}$ at every time $t$.
When the agent takes action $a_t\in \mathcal{A}$, the transition function $\mathcal{P}(\cdot)$ of the environment decides the next state $s_{t+1}=\mathcal{P}(s_t, a_t)$.
Then the environment gives the reward $r_{t+1}=\mathcal{R}(s_t, a_t)$ to the agent.
At every time $t$, the agent observes $s_t$ and decides the action $a_t\sim\pi(s_t)$ according to the policy $\pi(\cdot)$.
In the finite horizon MDP, value $V_\pi$ of the policy $\pi$ is defined as $V_\pi(s_t)=r_{t+1} + \gamma r_{t+2} + \dots + \gamma^{T-t-1} r_T$, \ie, the sum of discounted cumulative rewards until the end of the episode $t=T$.
The objective of RL is to find the optimal policy $\pi^*=\underset{\pi}{\mathrm{argmax}}\,V_\pi(s)$.

\bigbreak\textbf{State and Transition.}
In constructing the collage MDP for our RL agents, we highlight the three key components: canvas $C_t$, target image $I$, and material $M_t$ as illustrated in Fig. \ref{fig:overview}.
The collage MDP is initialized with white canvas $C_0$, target image $I\sim\mathcal{I}$, and the initial material image $M_0\sim\mathcal{M}$, where $\mathcal{I}$, and $\mathcal{M}$ are respectively the sets of target images and material images.
If the agent takes the action $a_t$ observing the canvas $C_t$, the material $M_t$ is cut and pasted onto the canvas, resulting in the next canvas $C_{t+1}=\delta(C_t, M_t, a_t)$.
The function $\delta$ is responsible for cut-and-paste, as an essential mechanism of collage state transition.
The next material $M_{t+1}$ is randomly sampled from $\mathcal{M}$.
Two additional components are included in the state.
One is CoordConv \cite{coordconv} $c$ that is a constant vector known to help DCNNs to boost its performance in coordinating tasks.
Another one is the remaining time $l_t=(T_M-t_M)/T_M$ about the number of material pieces that need to be pasted, where the constant $T_M$ is the total number of pieces to paste, and $t_M$ is the number of pieces already pasted before $t$.
In summary, the observable state for the agent is $s_t=(C_t,I,M_t,l_t,c)$ at time $t$.
The overall transition function can be represented using $\mathcal{P}$ as $s_{t+1}=\mathcal{P}(s_t,a_t)=(\delta(C_t,M_t,a_t),I,M_{t+1},(T_M-t_M)/T_M,c)$ where $M_{t+1}\sim\mathcal{M}$.

\bigbreak\textbf{Action Design.}
Each parameter of the agent's action determines the strategy for cutting and pasting the given material.
We define the action as a twelve-dimensional vector $a=\langle x_{cut},y_{cut},w,h,p_1,p_2,p_3,p_4,x_{glue},y_{glue},\theta,\upsilon\rangle$.
When cutting the material, the location, size, and shape are determined for a rectangle by center coordinates $x_{cut},y_{cut}$, width $w$, and height $h$.
Then the rectangle is divided into a quadrilateral scrap, following four points on each side based on the point ratio $p_{1:4}$.
When pasting the piece, the pasting location $x_{glue},y_{glue}$ and rotation $\theta$ are determined.
The material acceptor $\upsilon\in[0,1]$ determines whether to use the given material.
The agent can deny a poor given material and request another one if $\upsilon<0.5$, and $t_M$ increases only when the agent accepts the material.
If the agent denies the material at $t$, a new material is given, but the canvas and remaining time do not change as $C_{t+1}=C_t$, $l_{t+1}=l_t$.
The MDP normally terminates when $t_M=T_M$, but if the agent has denied too many given materials by $t=T_{max}$, the MDP terminates even if $t_M<T_M$.

\begin{figure}[t]
\begin{center}
\includegraphics[width=1\linewidth]{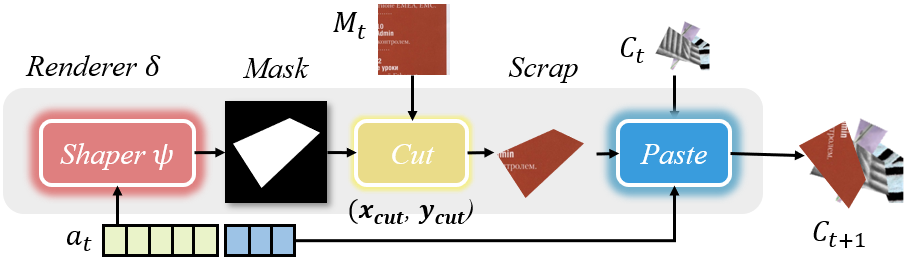}
\end{center}
\caption{The differentiable rendering process in our collage MDP. The actions determining the cutting shapes are input into the pre-trained shaper network $\psi$. The resulting mask is then used to cut the material, generating a scrap to be pasted onto the canvas.}
\label{fig:renderer}
\vspace{-4mm}  
\end{figure}

\bigbreak\textbf{Differentiable Collage.}
Since our training algorithm requires a differentiable $\mathcal{P}(s, a)$ about the action $a$, $\delta(C,M,a)$ should be differentiable.
To ensure the function $\delta$ is differentiable, all operations involved in the process, from selecting an action to generating the next canvas, must also be differentiable.
As in Fig. \ref{fig:renderer}, we generate a $mask$ that specifies where the location of the cutting region and shape of the material, but the process of generating $mask$ is typically non-differentiable.
To overcome this challenge, we use a pre-trained shaping network $\psi$ that is capable of generating differentiable $mask$ images. Additionally, the material acceptor $\upsilon$ is a discrete factor that is also non-differentiable.
We address this issue by including $\upsilon$ in the shaping network $\psi$, which has been pre-trained to output an all-zero mask when the material is denied.
To perform the material pasting, we use a differentiable transformation operation from Kornia \cite{kornia} that allows us to translate and rotate the material piece as needed. The new canvas is generated by adding the material piece to the current canvas, while excluding the $mask$ region.
Through these modifications, we have achieved a fully differentiable series of connections that allows for the implementation of $\delta$.

\bigbreak\textbf{Reward Function.} \label{chapter:reward_function}
The main goal of collage transfer is to create a final canvas $C_T$ that resembles the target image $I$ as a collage artwork.
As the style emerges naturally during the collage creation process, we designed a reward system that increases as the content of $C_T$ becomes more similar to $I$.
To measure the similarity between $C_T$ and $I$, we used a reward scheme proposed in \cite{learning2paint}, which calculates the amount of similarity change as $r_t = sim(C_{t-1}, I) - sim(C_t, I)$.
In this study, the discriminator in the Wasserstein GAN with gradient penalty (WGAN-GP) \cite{wgan-gp} was used for $sim$.
The $sim$ discriminator is trained simultaneously with the collage agent to discriminate the canvases-in-progress and the target images.
This dynamic reward system takes into account the agent's learning progress, resulting in better performance than using a constant reward, such as mean-squared error (MSE) \cite{learning2paint}.
In addition, we add step penalty -1 to the reward at every step to prevent the agent from neglecting materials.

\subsection{Training}

\textbf{Model-based SAC.}
We propose a modified version of model-free SAC \cite{sac}, called model-based Soft Actor-Critic (MB-SAC), to relieve the burden of learning collage dynamics in model-free RL.
SAC is based on the maximum entropy RL framework \cite{maxentropyrl} and adopts a stochastic policy approximating the optimal action distribution. Unlike traditional RL, the objective of the SAC policy update is not only to maximize the rewards but also to maximize the entropy of the policy. 

The difference between the value functions of traditional RL and SAC is described as
\begin{gather}
    V^{RL}_\pi(s_t)=\mathbb{E}_{a_t\sim\pi}\left[Q(s_t,a_t) \right] \label{eq:vrl}, \\
    V^{SAC}_\pi(s_t)=\mathbb{E}_{a_t\sim\pi}\left[Q(s_t,a_t)-\log\pi(a_t|s_t) \right] \label{eq:vsac}.
\end{gather}
This ensures that the policy being trained behaves as randomly as possible while maximizing exploration, improving learning efficiency, and effectively finding improved behaviors.
As an off-policy policy gradient method with an actor-critic structure, Model-free SAC utilizes past experience for training.
When the past experience is stored in $\mathcal{D}$, the objectives of SAC policy evaluation and policy improvement are as follows \cite{sac2}.
\begin{multline}
    J_Q=\mathbb{E}_{(s_t,a_t)\sim\mathcal{D}}[\tfrac{1}{2}(Q(s_t,a_t) \\
    -(r(s_t,a_t)+\gamma\mathbb{E}_{s_{t+1}\sim\mathcal{P}}\left[V(s_{t+1})\right]))^2] \label{eq:sac_q_objective}
\end{multline}
\begin{gather}
    J_\pi=\mathbb{E}_{s_t\sim\mathcal{D}}\left[\mathbb{E}_{a_t\sim\pi}\left[\alpha\log(\pi(a_t|s_t))-Q(s_t,a_t)\right]\right] \label{eq:sac_policy_objective}
\end{gather}

Model-based RL (MBRL) \cite{mbrl} is an RL extension that can be used when the transition $\mathcal{P}$ and the reward function $\mathcal{R}$ are known or approximated.
If the model is correct, the training agent can leverage this knowledge about transitions and rewards to alleviate the training burden.
In collage MDP, the $\mathcal{P}$ is known as a differentiable operator $\delta$.
The reward function is also known as described in \ref{chapter:reward_function}. Thus, we can benefit from the model-based architecture.
Zhou \etal \cite{learning2paint} modified Deep Deterministic Policy Gradient (DDPG) \cite{ddpg} to a model-based version using the objective \eqref{eq:vrl}, but we use the alternative objective \eqref{eq:vsac} to enhance exploration and performance of the agent.
For our MB-SAC formulation, we use the equation \eqref{eq:vsac} and its interchangeable relationship
\begin{gather}
    Q(s_t,a_t)=r(s_t,a_t)+\mathbb{E}_{s_{t+1}\sim\mathcal{P}}\left[V(s_{t+1})\right], \label{eq:qv_relation}
\end{gather}
to reformulate the objectives \eqref{eq:sac_q_objective} and \eqref{eq:sac_policy_objective} as the following objectives:
\begin{multline}
    J_V=\mathbb{E}_{(s_t,a_t)\sim\mathcal{D}}[\tfrac{1}{2}(V(s_{t+1})-\mathbb{E}_{a_{t+1}\sim\pi}[r(s_{t+1},a_{t+1}) \\
    +\gamma\mathbb{E}_{s_{t+2}\sim\mathcal{P}}[V(s_{t+2})]-\alpha\log(\pi(a_{t+1}|s_{t+1}))])^2] ,\label{eq:mbsac_q_objective}
\end{multline}
\begin{multline}
    J_\pi=\mathbb{E}_{s_t\sim\mathcal{D}}[\mathbb{E}_{a_t\sim\pi}[\alpha\log(\pi(a_t|s_t)) \\
    -(r(s_t,a_t)+\mathbb{E}_{s_{t+1}\sim\mathcal{P}}[V(s_{t+1})])]]. \label{eq:mbsac_policy_objective}
\end{multline}
Since $\mathcal{P}$ and $\mathcal{R}$ are differentiable, the above objectives allow policy gradients on the known transition and reward dynamics.

It should be noted that the transition model is not completely accurate due to a small error in $\psi$, and the agent is unaware of the upcoming material.
However, for uncertain future events, the expectations in equations \eqref{eq:mbsac_q_objective} and \eqref{eq:mbsac_policy_objective} combine the potential outcomes of $\mathcal{P}$, and therefore do not necessarily require knowledge of the upcoming material.
Hence, the agent simply samples a new random material using the transition model to predict the next state.

\bigbreak\textbf{Training Scheme.}
The agent learns to make a collage on the defined MDP, over a time range from $t=0$ to at least $t=T_M$ (up to $t=T_{max}$) as one episode following the objectives \eqref{eq:mbsac_q_objective} and \eqref{eq:mbsac_policy_objective}.
As the agent iterates through episodes, it gains experience in various state combinations $(C, M, I, t)$.
At the end of each episode, the RL and $sim$ are updated together.
This creates a virtuous cycle as $sim$ gets tighter and tighter and the policy gets more and more accurate.

\subsection{Advanced Techniques}
\textbf{Active Material Selection.}
During training, the agent makes binary decisions to accept or deny randomly given materials.
However, humans actively select the best choice among multiple materials.
We utilize the trained action-value function $Q$ denoted in \eqref{eq:qv_relation} without needing an additional material selector to achieve active material selection like a human.
If $Q$ is accurate, then the optimal material for a given state is the one that maximizes $Q$, since $Q$ represents the value of a state-action pair.
The agent trains $V$ instead of $Q$, but $Q$ can be recovered using $V$ and the reward function $\mathcal{R}$, as shown in \eqref{eq:qv_relation}.
Therefore, the optimal material $m^*_t$ from the set $\mathcal{M}$ can be selected using the following equation:
\begin{multline}
    m^*_t=\underset{m}{\mathrm{argmax}}\left(r(s_t,a_t)+\gamma V(s_{t+1}) \right),\,m\in\mathcal{M}, \\ \,a_t=\mathbb{E}_{s_t\sim\mathcal{P}}\left[\pi(s_t)\right],\,s_{t+1}=\mathcal{P}(s_t,a_t).
\end{multline}
It is worth noting that the agent can still deny the material even if it is optimal, especially if the canvas is already mature enough.
In practice, the agent finds $m^*_t$ from a subset of $\mathcal{M}$ to avoid high computational costs.

\begin{figure*}[t]
\centering
\begin{subfigure}{0.3\textwidth}
    \includegraphics[width=\textwidth]{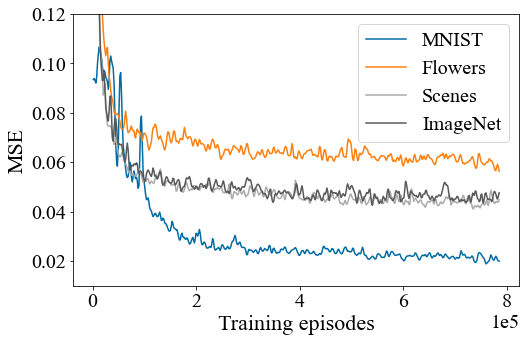}
    \caption{}
    \label{fig:ssc_learning_curve_a}
\end{subfigure}
\hfill
\begin{subfigure}{0.3\textwidth}
    \includegraphics[width=\textwidth]{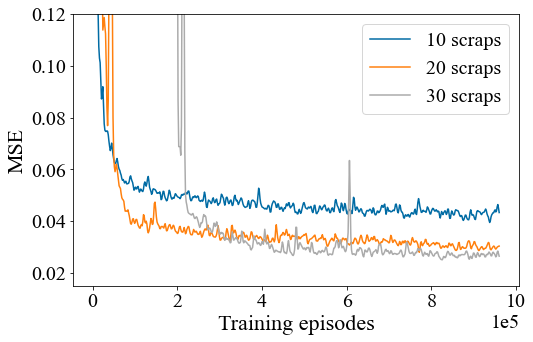}
    \caption{}
    \label{fig:ssc_learning_curve_b}
\end{subfigure}
\hfill
\begin{subfigure}{0.3\textwidth}
    \includegraphics[width=\textwidth]{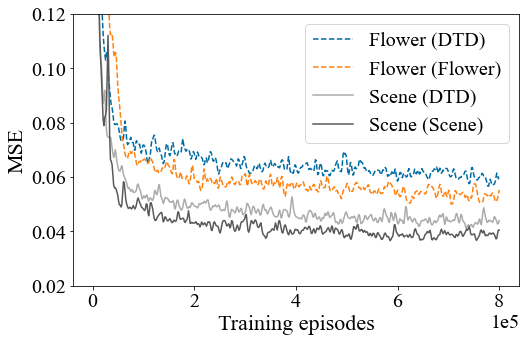}
    \caption{}
    \label{fig:ssc_learning_curve_c}
\end{subfigure}
\caption{The learning curves of the model-based collage RL: (a) the learning curves of different target image domains using DTD, (b) the learning curves of agents trained to paste 10, 20, and 30 scraps using DTD, and (c) the learning curves of each agent using DTD and using materials from the same target domain.}
\label{fig:ssc_learning_curve}
\vspace{-2mm}  
\end{figure*}

\bigbreak\textbf{Multi-Scale Collage.}
We employ a multi-scale collage similar to the underpainting process to achieve a coarse-to-fine strategy for any size of an image.
We define the multi-scale sequence $\mathcal{U}=(u_1,u_2,\dots,u_n),$ where $u_1>u_2>\dots>u_n$ and $u,n\in\mathbb{N}$ for the target image $I$ with size $W\times H$.
Then $I$ is divided into smaller images using a sliding window with a window size of $u\times u$ and stride of $\lceil\rho\cdot u\rceil$, where $\rho$ is the stride ratio.
The number of divided target image regions on scale $u$ is calculated as $k(u)=(\lceil(W-u)/\rho\rceil+1)(\lceil(H-u)/\rho\rceil+1)$.
Finally, we acquire the set of divided target images $\mathbf{I}(u)$ and the set of corresponding initial divided canvases $\mathbf{C}(u)_{t=0}$ for every scale $u$.

\begin{gather}
    \mathbf{I}(u)=(I(u)^1,I(u)^2,\dots,I(u)^{k(u)}) \label{eq:divided_goals} \\
    \mathbf{C}(u)_{t=0}=(C(u)^1_{t=0},C(u)^2_{t=0},\dots,C(u)^{k(u)}_{t=0})
\end{gather}
Each divided target image is resized uniformly to the trained network’s fixed input size during multi-scale collage.
Therefore, the agent at earlier scales observes more broad area than at later scales but with higher information loss due to the resizing gap.
We define ``1-cycle" as one step of progress at each pair $(I(u)^i, C(u)^i_t)$ in a given scale $u$.
A multi-scale collage takes $K$-cycles for each scale $u$, in the order $u_1,u_2,\dots,u_n$, to create the final collage.
The initial canvases on the first scale are all white, but the initial canvases of scales after the second inherit the final canvases of the previous scales.

\bigbreak\textbf{Complexity-Aware Multi-Scale Collage.}
When creating a multi-scale collage, we utilize the insight that areas of a drawing with higher complexity require more intricate touches.
We measure image complexity $Co(I)$ of the image $I$ using an image gradient-based complexity measure introduced in \cite{aestheticmeasures}. 
We use the Sobel filter for the image gradient.
The complexity is high when high color changes like edges are prevalent.
Using this complexity measure, a set of complexities $\mathbf{Co}(I)_u$ of divided target images in \eqref{eq:divided_goals} can be calculated for every $u$.
\begin{gather}
    \mathbf{Co}(I)_u=(Co(I(u)^1),Co(I(u)^2),\dots,Co(I(u)^{k(u)})) 
\end{gather}
Assuming the divided image complexities follow a normal distribution, we have the standardized complexities $\mathbf{Co}_Z(I)_u$ by the mean of complexity $\mu$ and the standard deviation $\sigma$ from $\mathbf{Co}(I)_u$.

Referring to cumulative probability $Co_p(I(u)^i)$ for each $z$-score in $\mathbf{Co}_Z(I)_u$, the number of scraps for each divided canvas can be assigned proportional to $Co_p$.
The scraps can be adjusted if the number of cycles $K$ for each divided canvas are adjusted.
Therefore, we applied monotonically increasing $K$ with respect to the $Co_p$ by the following equation
\begin{equation}
    K(Co_p) = K_{max}\cdot Co_p^\tau,
\end{equation}
where $K_{max}$ is the maximum $K$ for each divided canvas and $\tau\in (0,\infty)$ determines the sensitivity.

\section{Experiments}
To validate the proposed neural collage transfer method, this section presents implementation details of the training process, a comprehensive analysis of experimental results, and a comparison of the outcomes with existing methods.

\subsection{Training}
\textbf{Implementation Details.}
For the backbones of actor and critic networks of the RL agents, ResNet-18 \cite{resnet} with weight normalization \cite{weightnorm} was adopted. 
Batch normalization \cite{batchnorm} was applied to the actor.
The discriminator used vanilla CNN with weight normalization, followed by global average pooling \cite{gap} at the end of the network.
Translated ReLU \cite{trelu} served as the activation function for all networks.
Prior to training, $\psi$ was pre-trained with pixel shuffle \cite{pixelshuffle} for 100k epochs, using pairs of random actions and corresponding masks with batch size 64.
During training, the differentiable $\delta$ was only used for MB-SAC updates, while a non-differentiable transition function was utilized for the real MDP step, which excluded any errors from $\psi$.
The replay memory size was set to 20K, and the agents were trained in 16 parallel environments.
At the end of each episode, 5 RL updates and one discriminator update were performed, with a batch size of 64.

\bigbreak\textbf{Datasets.}
For our experiments, we utilized various domain images, including MNIST \cite{mnist}, Flower \cite{flowers}, Scene\footnote{https://www.kaggle.com/datasets/puneet6060/intel-image-classification}, and ImageNet \cite{imagenet} as target image domains.
For material images, we utilized Describable Textures Dataset (DTD) \cite{dtd}, Times\footnote{https://www.kaggle.com/datasets/thegupta/time-magazine-part-1-1923-to-1930}, and Newspaper \cite{newspaper}.
We adopted DTD as the primary material images for training due to its diverse textures and colors.
Newspaper and Times were used as materials for multi-scale collage, as they are common types used in general collage.
The data was split into training and evaluation images, ensuring that the agent encountered new images during evaluations.

\begin{figure}[t]
\begin{center}
\includegraphics[width=1.0\linewidth]{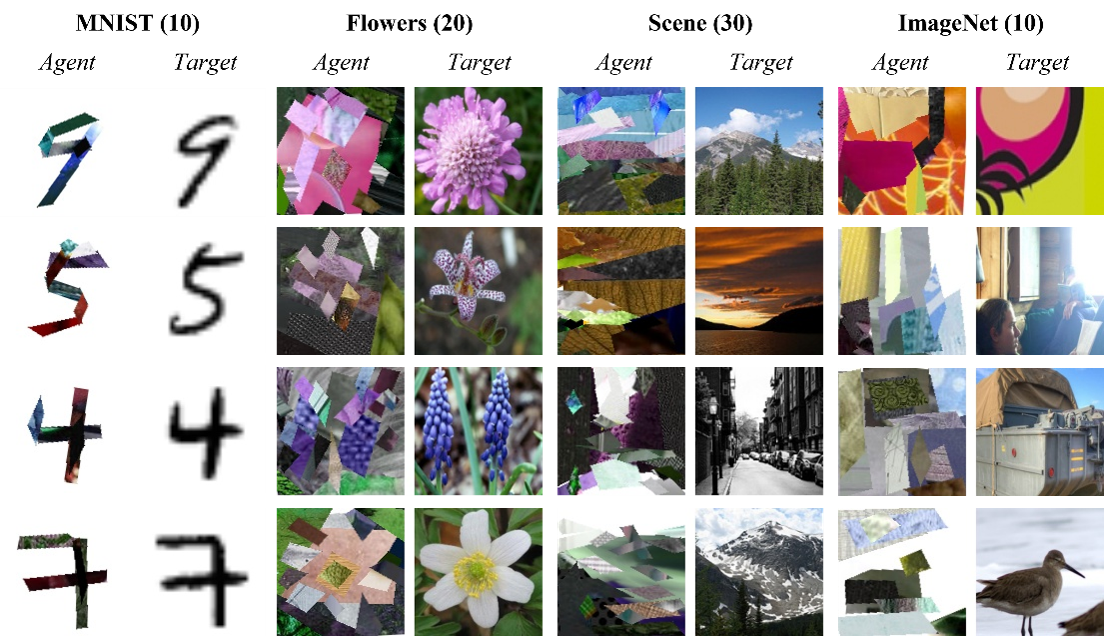}
\end{center}
\caption{Single-scale collage results using DTD.}
\label{fig:ssc_dtd}
\vspace{-4mm}  
\end{figure}

\bigbreak\textbf{Single-Scale Collage.}
To investigate the pure behavior of our agents, we demonstrated the agent’s RL-based learning process and results on the fixed image size $128\times128$ without considering image division or complexities.
Fig. \ref{fig:ssc_learning_curve_a} shows a learning curve of each agent trained using DTD on four target image domains.
 The MSE between the target image and the collage generated by the agent decreased as the training episode progressed.
However, owing to image structure and domain variations, the range of MSE differed across the target images.
Fig. \ref{fig:ssc_dtd} illustrates each agent's collage results using DTD, with a different number of scraps applied to each domain, as indicated in parentheses. 
For MNIST, the numbers in results are clearly recognizable, being placed with appropriate scrap lengths and sizes.
For Flower and Scene, the scraps are correctly pasted considering their shapes and colors, even though the materials have different properties than the target image domain.
Notably, for the result in the third column in Flower, the agent could even consider textures if the materials permitted.
As ImageNet contains an extensive variety of images, the agent focused on reproducing general colors and shapes, rather than domain-specific collages.
Overall, the agent selected and arranged the materials at the appropriate level, resulting in collages that closely resemble the target images at a glance, while each pasted scrap retains its independence when viewed up close.

\begin{figure}[t]
\begin{center}
\includegraphics[width=1.0\linewidth]{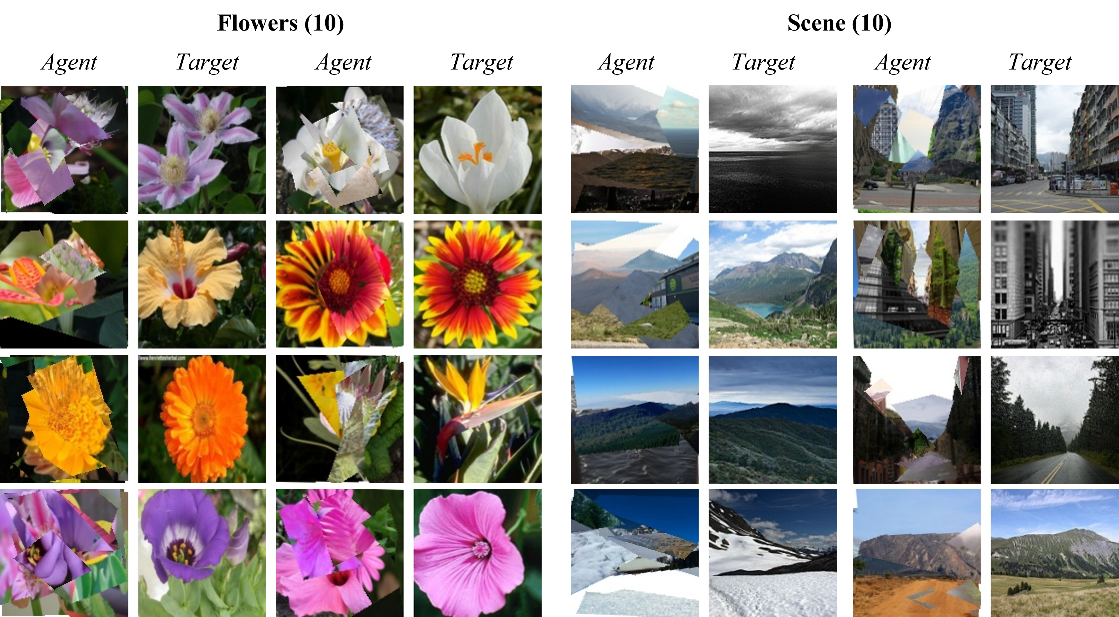}
\end{center}
\caption{Single-scale collage results using materials from the same target domain.}
\label{fig:ssc_self}
\vspace{-4mm}  
\end{figure}

Fig. \ref{fig:ssc_learning_curve_b} illustrates the learning curve of each agent trained to paste 10, 20, and 30 scraps using DTD for the target images from Scene.
As the number of scraps increases, the MSE decreases because more room to refine representation is available when more scraps are given.
However, it is worth noting that using many scraps does not always lead to high-quality collage; as too many scraps may harm the nature of the collage.
It depends on the user's preferred style.
Fig. \ref{fig:ssc_learning_curve_c} displays the differences in MSE between using DTD and the same materials as the corresponding target images.
Because images in the same domain have similar shapes and colors, a lower MSE was achieved when the material image domain was the same as the target image domain.
Resulting images of these cases are depicted in Fig. \ref{fig:ssc_self}.
Although the agents learned without labels, they used similar types of scraps with a target area and sometimes placed scraps with similar shapes and colors in different classes.

\bigbreak\textbf{Complexity-aware Multi-scale Collage.} \label{chapter:multi-scale_exp}
To validate our proposed multi-scale collage approach with complexity comprehension, we trained the agent on ImageNet as the target image domain and DTD.
The agent was trained to paste 10 scraps, and the input image size was $64\times64$.
In the single-scale collage, the agent tended to select smaller scraps as $t_m$ increased.
To encourage the agent to use more detailed scraps, we fixed $t_m=9$ (not always good).
For the hyper-parameters, we used $\mathcal{U}=\{512,256,128,64,32\}$, $\rho=0.5$, $K_{max}=8$ and $\tau=1$.
Newspapers and Times materials are used with active material selection techniques.
The final results and the results for each scale are shown in Fig. 1.
As illustrated below, We found that as the scale increased, the collage became more abstract and unique in style but with higher distortion.
On the other hand, smaller scales focused more on precise representation but lost some unique collage style.
Overall, the number of scraps pasted on the canvas differed depending on the target image's partial complexity.

\begin{figure*}[t]
\begin{center}
\includegraphics[width=1\linewidth]{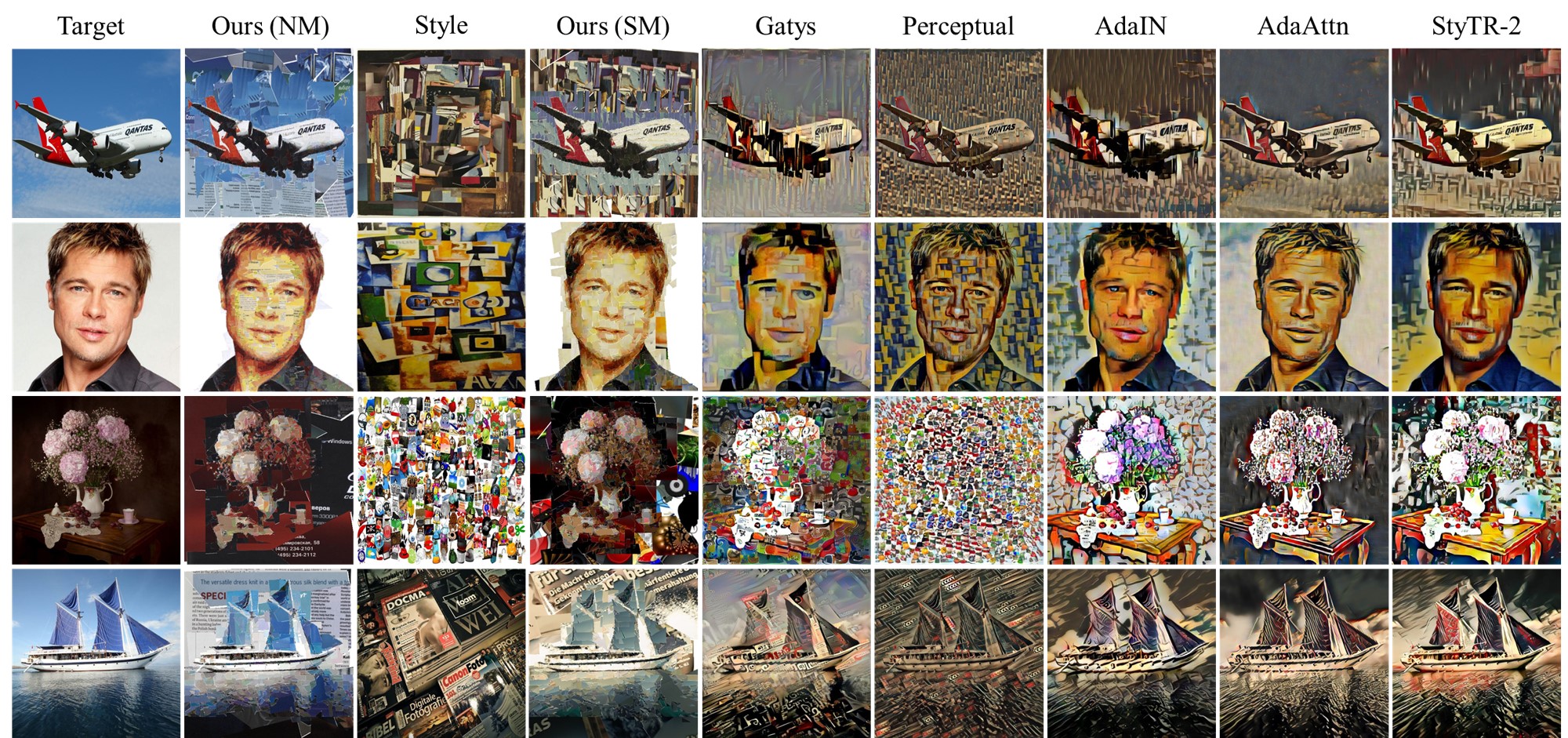}
\end{center}
\caption{Comparison of NST results using a given collage style images and the results from the proposed method. Ours (NM) used numerous other materials, while Ours (SM) solely used the given collage-style image as materials.}
\label{fig:nst_results}
\vspace{-4mm}  
\end{figure*}

\begin{table}[t]
\centering
\setlength{\tabcolsep}{1pt}
\renewcommand{\arraystretch}{1.3}
\resizebox{\columnwidth}{!}
{\scriptsize
\begin{tabular}{ccccccccc}
\Xhline{2.5\arrayrulewidth}
\multirow{2}{*}{\textit{Methods}} & \multicolumn{3}{c}{\textit{CLIP score} $\uparrow$ \cite{radford2021learning}} &  & \textit{CLIP vote} $\uparrow$ &  &  \textit{LPIPS} \cite{lpips} $\downarrow$ \\ \cline{2-4} \cline{6-6} \cline{8-8}
                            & \textit{content}            & \textit{human}              & \textit{collage}            & & \textit{collage} & & VGG  \\ \hline
\textit{Target}                  & 0.276 {\tiny$\pm$ 0.027}        & 0.213 {\tiny$\pm$ 0.018}        & 0.200 {\tiny$\pm$ 0.017}             & & 0.633  & & -  \\
AdaAttn \cite{adaattn}      & 0.278 {\tiny$\pm$ 0.021}         & 0.247 {\tiny$\pm$ 0.018}          & 0.241 {\tiny$\pm$ 0.010}            & & 0.027  & & 0.597 {\tiny$\pm$ 0.103}  \\
Adain \cite{adain}          & 0.251 {\tiny$\pm$ 0.019}         & 0.239 {\tiny$\pm$ 0.010}           & 0.236 {\tiny$\pm$ 0.008}           & & 0.017  & & 0.662 {\tiny$\pm$ 0.103 } \\
Gatys \cite{gatys}          & 0.226 {\tiny$\pm$ 0.013}         & 0.260 {\tiny$\pm$ 0.006}           & 0.250 {\tiny$\pm$ 0.006}            & & 0.290   & & 0.708 {\tiny$\pm$ 0.098}  \\
Perceptual \cite{perceptual}& 0.239 {\tiny$\pm$ 0.019}         & 0.246 {\tiny$\pm$ 0.006}          & 0.234 {\tiny$\pm$ 0.007}           & & 0.307  & & 0.722 {\tiny$\pm$ 0.117} \\
StyTR-2 \cite{stytr2}       &0.261 {\tiny$\pm$ 0.023}          &0.238 {\tiny$\pm$ 0.010}            &0.235 {\tiny$\pm$ 0.009}            & & 0.027  & & 0.613 {\tiny$\pm$ 0.115}  \\ \hline
Ours (32)                   &\textbf{0.280 {\tiny$\pm$ 0.026}} & 0.262 {\tiny$\pm$ 0.017}          & 0.281 {\tiny$\pm$ 0.020}            & & 0.100  & & \textbf{0.510 {\tiny$\pm$ 0.111}} \\
Ours (64)                   & 0.262 {\tiny$\pm$ 0.028}         & 0.272 {\tiny$\pm$ 0.020}          & 0.259 {\tiny$\pm$ 0.015}           & & 0.667  & & 0.565 {\tiny$\pm$ 0.112} \\
Ours (128)                  & 0.225 {\tiny$\pm$ 0.023}         &\textbf{0.288 {\tiny$\pm$ 0.015}}  &\textbf{0.272 {\tiny$\pm$ 0.016}}  & &\textbf{1.000} & & 0.610 {\tiny$\pm$ 0.115} \\ \Xhline{2.5\arrayrulewidth} \\
\end{tabular}}
\caption{Quantitative results in terms of CLIP score and LPIPS for the generated images from different NST methods and ours.}
\label{tab:comparison}
\vspace{-9mm}  
\end{table}

\subsection{Comparison with NST}
To evaluate our method's content maintenance and collage style agreement comparable to NST, quantitative and qualitative comparisons with the five NST methods \cite{gatys,adain,adaattn,perceptual,p2gan} were conducted.
For the quantitative comparison, metric-based evaluation and user study were taken.

For the metric-based evaluation, we prepared 30 target images for each method.
Using these 30 target images, we collected 30 pairs of (target image, generated image) for our multi-scale method.
The 30 pairs for each NST method were collected using style images consisting of five human-made collage images and five cluttered paper images.
We adopted CLIP score \cite{radford2021learning} for both the content maintenance metric and style agreement metric and adopted LPIPS \cite{lpips} only for the content maintenance metric.
Here, the CLIP score is the cosine similarity between the CLIP representation of the image and text.
We considered CLIP as a precise and reproducible protocol and even objective since CLIP was trained with about 400M human-labeled data.

Table \ref{tab:comparison} demonstrates the metric-based evaluation.
Each value in the table is the means and standard deviations for the 30 pairs.
In \textit{CLIP score} experiment, the methods are evaluated using the following texts: (\textit{content}) ``a \textit{content name}", (\textit{human}) ``a human-made collage image'', and (\textit{collage}) ``a collage artwork", where \textit{content name} is the name of each target image (\eg, a bird).
We assumed the content score as the CLIP score of the text-image pair (\textit{content}, generated image) and the style score as the CLIP score of the pair (\textit{human/collage}, generated image).
The rows under \textit{CLIP score} show that our generated images achieved a relatively high content and style score.
However, the content score significantly decreased as the scale grew, indicating that larger scales emphasized collage style over content.
\begin{table}[t]
\resizebox{\columnwidth}{!}{%
\begin{tabular}{cccccc}
\Xhline{3\arrayrulewidth}\\[-2.5ex]
                 & \multicolumn{1}{c}{\textit{style} $\uparrow$}         & \multicolumn{1}{c}{\textit{content} $\uparrow$}      & \multicolumn{1}{c}{\textit{aesthetic} $\uparrow$}     & \multicolumn{1}{c}{\textit{ai} $\downarrow$}           & \multicolumn{1}{c}{\textit{human} $\uparrow$}    \\[-2.5ex]     \\ \hline 
\\[-2.5ex]
AdaAttn          & 18.3\scriptsize $\pm$ 9.2  & 38.4 \scriptsize$\pm$ 14.8 & 30.0 \scriptsize$\pm$ 10.3   & 34.2 \scriptsize$\pm$ 6.5 & 31.6 \scriptsize$\pm$ 11.2 \\
AdaIN            & 24.9 \scriptsize$\pm$ 8.4  & 21.1 \scriptsize$\pm$ 8.2  & 26.7 \scriptsize$\pm$ 1.5  & 35.8 \scriptsize$\pm$ 5.1 & 24.3 \scriptsize$\pm$ 5.8  \\
Gatys            & 35.5 \scriptsize$\pm$ 5.4  & 17.4 \scriptsize$\pm$ 6.5  & 26.6 \scriptsize$\pm$ 10.2 & 35.6 \scriptsize$\pm$ 6.2 & 27.4 \scriptsize$\pm$ 9.4  \\
Perceptual & 26.2 \scriptsize$\pm$ 8.2  & 6.7 \scriptsize$\pm$ 3.4   & 25.4 \scriptsize$\pm$ 5.0    & 37.0 \scriptsize$\pm$ 7.9   & 25.1\scriptsize$\pm$ 7.6  \\
StyTR-2          & 30.5 \scriptsize$\pm$ 3.2  & 44.1 \scriptsize$\pm$ 20.3 & 41.4 \scriptsize$\pm$ 8.5  & 30.4 \scriptsize$\pm$ 8.1 & 31.5 \scriptsize$\pm$ 9.1  \\
Ours             & \textbf{64.6 \scriptsize$\pm$ 13.2} & \textbf{72.2 \scriptsize$\pm$ 13.3} & \textbf{49.9 \scriptsize$\pm$ 8.9}    & \textbf{27.0 \scriptsize$\pm$ 6.0}   & \textbf{60.2 \scriptsize$\pm$ 13.4} \\[-2.5ex]  \\ \hline \\[-2.5ex]
Ours (32)         & 51.8 \scriptsize$\pm$ 4.5  & \textbf{92.8 \scriptsize$\pm$ 4.7}  & \textbf{73.4 \scriptsize$\pm$ 5.3}  & -                                & -                                 \\
Ours (64)         & 72.5 \scriptsize$\pm$ 0.6  & 67.0 \scriptsize$\pm$ 0.8    & 70.4 \scriptsize$\pm$ 2.8    & -                                & -                                 \\
Ours (128)        & \textbf{75.7 \scriptsize$\pm$ 5.1}  & 40.2 \scriptsize$\pm$ 3.8  & 56.2 \scriptsize$\pm$ 8.1  & -                                & -                                 \\ \\[-2.5ex] \Xhline{3\arrayrulewidth} \\ \vspace{-7mm}
\end{tabular}%
}
\caption{Quantitative analysis from user study.}
\label{tab:survey}
\vspace{-8mm}  
\end{table}
The style scores of our generated images were always the highest.
In \textit{CLIP vote} experiment, CLIP predicted the probabilities of how much the generated image matches with the following two texts: (\textit{collage}) ``human-made collage", (\textit{nst}) and ``neural style transferred artwork."
This was intended to investigate if CLIP can distinguish between the NST results and our results based on the semantics of ``collage'' and ``neural style transferred."
The rows under \textit{CLIP vote} show that our generated images at larger scales gained the highest probability on \textit{collage} compared to \textit{nst}, but it decreased at scale 32, showing the scale-dependent style-content trade-off of our method.

Further content-focused evaluation using LPIPS with VGG \cite{vgg} was conducted.
According to the values in rows under \textit{LPIPS}, our approach achieved a relatively low LPIPS distance, confirming that the proposed method can stably maintain the content of the target image.
Since NST treats the target image's color as style, LPIPS distance was relatively low for our approach, which treats the color as content.
These results make sense since NST does not use various materials and does not actually create a collage.

We also conducted a user study on 111 non-conflicting respondents, as in Table \ref{tab:survey}.
Each user responded to the three to four question sets.
Each set firstly displayed the user a target image and the generated images from NSTs and our method.
Then it required the user to rank the top three images for the following five questions: ``Which one is the most ..." (\textit{style}) ``collage-style?" (\textit{content}) ``content-preserving?" (\textit{aesthetic}) ``aesthetically appealing?" (\textit{ai}) ``AI-generated?" (\textit{human}) ``human-created?".
The scores were averaged, with weights of 3, 2, and 1 for each rank.
Surprisingly, the result aligns well with the metric-based evaluation.
Each user also responded to the two additional question sets to rank our generated images at scales 128, 64, and 32 for the questions \textit{style}, \textit{aesthetic}, and \textit{human}.
Once again, the result supports findings from the metric-based evaluation.

Fig. \ref{fig:nst_results} illustrates the transfer examples of our and NST methods.
Ours (NM) refers to \textit{numerous materials}, which means that other images were used as materials, regardless of the style image.
Ours (SM) refers to \textit{single material}, meaning only the given style image was used as material.
Although NST produced aesthetically appealing transferred results, it did not achieve the intended collage style.
Ours (SM) constructed the collage artwork from white canvas, even though only the single style image was given as a material.
While NST refers to a single style image and follows the colors in the style image, our method can select materials with proper colors from other image sources to construct content-color-maintaining collages if materials are abundant, as illustrated in the results of Ours (NM).

\begin{table}[t]
\centering
\setlength{\tabcolsep}{9pt}
\resizebox{\columnwidth}{!}
{
\begin{tabular}{lccc}
\Xhline{3\arrayrulewidth} \\[-2.5ex]\
 & MSE $\downarrow$ & PSNR $\uparrow$ & SSIM $\uparrow$ \\ \hline \\[-2.5ex]
\textit{RL algorithms} & & &  \\ 
DDPG \cite{ddpg} & 0.305 {\scriptsize$\pm$ 0.144} & 6.092 {\scriptsize$\pm$ 3.919} & 0.213 {\scriptsize$\pm$ 0.034} \\
SAC \cite{sac} & 0.042 {\scriptsize$\pm$ 0.002} & 13.78 {\scriptsize$\pm$ 0.285} & 0.191 {\scriptsize$\pm$ 0.013} \\
MB-DDPG \cite{learning2paint} & 0.039 {\scriptsize$\pm$ 0.004} & 14.06 {\scriptsize$\pm$ 0.514} & 0.204 {\scriptsize$\pm$ 0.016} \\ 
MB-SAC (ours) & \textbf{0.031 {\scriptsize$\pm$ 0.001}} & \textbf{15.09 {\scriptsize$\pm$ 0.219}} & \textbf{0.220 {\scriptsize$\pm$ 0.004}} \\  \\[-2.5ex] 
\hline
\textit{Rewards} & & &  \\
MSE & 0.122 {\scriptsize$\pm$ 0.025} & 9.191 {\scriptsize$\pm$ 0.893} & 0.101 {\scriptsize$\pm$ 0.018} \\
WGAN-GP \cite{wgan-gp} & \textbf{0.031 {\scriptsize$\pm$ 0.001}} & \textbf{15.09 {\scriptsize$\pm$ 0.219}} & \textbf{0.220 {\scriptsize$\pm$ 0.004}} \\
SNGAN \cite{sngan} & 0.040 {\scriptsize$\pm$ 0.002} & 13.96 {\scriptsize$\pm$ 0.256} & 0.205 {\scriptsize$\pm$ 0.014} \\
GNGAN \cite{gngan} & 0.084 {\scriptsize$\pm$ 0.035} & 11.06 {\scriptsize$\pm$ 1.933} & 0.144 {\scriptsize$\pm$ 0.038} \\ \\[-2.5ex]\Xhline{3\arrayrulewidth} \\
\end{tabular}
}
\caption{The results of ablations under the different RL algorithms and reward components.}
\label{tab_reward}
\vspace{-8mm}  
\end{table}

\subsection{Ablation Studies}
Ablation studies were conducted to investigate the effects of the main components of training.
In this study, agents were trained using Scene as target images and DTD as materials.
The agents were trained with 800K episodes for five seeds using input image size $64\times64$.

\bigbreak\textbf{RL Algorithm.}
To confirm if the proposed MB-SAC outperforms existing algorithms, we evaluated four RL algorithms, model-free DDPG \cite{ddpg}, model-free SAC \cite{sac}, MB-DDPG \cite{learning2paint}, and MB-SAC.
Three evaluation metrics, MSE, PSNR, and SSIM, are used as image distance measure.
The resulting values are described in table \ref{tab_reward}.
The model-free DDPG almost failed to learn, while the model-free SAC succeeded.
DDPG's failure was likely due to the high complexity and uncertainty of the collage procedure, which SAC was able to withstand.
Model-based approaches show better performance than model-free as the model-based approach successfully reduced the learning burden of the transition and reward of the MDP.
As expected, our MB-SAC achieved the best performance by leveraging the strengths of both SAC and model-based approach.

\bigbreak\textbf{Reward Function.}
Since the reward is a vital signal for RL, we investigated alternative discriminators \cite{sngan, gngan} proposed after WGAN-GP as demonstrated in table \ref{tab_reward}.
Among them, WGAN-GP is still the most potent signal for our RL agent, though SNGAN \cite{sngan} and GNGAN \cite{gngan} are reported to perform better in general GAN training.
Both methods stabilize general GAN training by limiting the output of the discriminator, but it seems that they distort the distance metric between images, making them less informative when used as rewards for RL.
Generally, a dense reward with a precise scale is considered helpful than a sparse reward; in that regard, WGAN-GP, which mostly maintains the \textit{Wasserestein} distance was the most informative discriminator for reward signals.
The MSE reward signal showed poor performance, even in the MSE evaluation.
It indicates that using the reward signal relative to the performance (\eg GANs) of the learning agent was critical in training.

\section{Conclusion and Future Works}
This paper presented a novel RL-based training architecture, MB-SAC, and complexity-aware multi-scale collage techniques for stroke-based collage transfer.
The experimental results demonstrated that our method enables the agents to autonomously learn to create collage with its distinct style without demonstration data.
The agents could handle target images of any size and generate sophisticated and aesthetically pleasing output while maintaining collage style and content.

Since this study represents the first exploration of neural collage transfer using materials as-is, several limitations provide ample opportunity for future improvements.
Following the proposed method, the stroke shape is limited to quadrilateral, so extending it to more unconstrained shapes is a promising future direction.
The stroke choice only considers color and shape, disregarding semantics.
Incorporating semantics or contexts of materials and targets would also be an interesting extension.
Adding custom reward factors to reflect intentional distortions or style variations is also possible.
For practicality, adding interactive commands to cater to the users' tastes would be a powerful use case.
Additionally, the proposed methods can be extended beyond collage to other art forms that may not align with traditional pixel-based or fixed-stroke-based approaches.

\bigbreak\textbf{Acknowledgments.}
This work was partly supported by the NRF of Korea (2021R1A2C1010970/20\%), %
and the IITP (2021-0-02068-AIHub/10\%, %
2021-0-01343-GSAI/10\%, %
2022-0-00951-LBA/30\%, %
2022-0-00953-PICA/30\%) %
grants funded by the Korean government.
We would like to thank Ane Kristine Espeseth and Min Whoo Lee for the English editing.

{\small
\bibliographystyle{ieee_fullname}
\bibliography{egbib}
}

\end{document}